\PassOptionsToPackage{para}{manyfoot}
\PassOptionsToPackage{hyperfootnotes=false}{hyperref}
\documentclass[acmsmall,screen,nonacm]{acmart}
\usepackage{tabularx}
\usepackage{xcolor}
\usepackage{pifont}
\usepackage{multirow}
\usepackage{enumitem}
\usepackage{subcaption}
\DeclareNewFootnote[para]{inline}
\makeatletter
\patchcmd{\@typeset@author@line}
  {\unskip, {\@currentaffiliations}}
  {\unskip\par\noindent{\@currentaffiliations}}
  {}{}
\patchcmd{\@mkauthorsaddresses}
  {\def\and{\unskip; \gdef\@ACM@institution@separator{, }}}
  {\def\and{\unskip; \gdef\@ACM@institution@separator{\quad}}}
  {}{}
\makeatother
\AtBeginDocument{}

\begin{document}

\title{AlphaTruck: A Platform for High-Speed Off-Road Autonomy on Unstructured Terrain}

\author{Rwik Rana$^{1}$}
\author{Adam J. Thorpe$^{1}$}
\author{Jesse Quattrociocchi$^{1,2}$}
\author{William Ward$^{1}$}
\author{\\Goktug Poyrazoglu$^{1}$}
\author{Adam Uccello$^{1,2}$}
\author{Christian Ellis$^{1,2}$}
\author{\\Rahul Moorthy Mahesh$^{1}$}
\author{Yukang Cao$^{1}$}
\author{Nathan Tsoi$^{1}$}
\author{Garrett Warnell$^{1,2}$}
\author{Volkan Isler$^{1}$}
\author{Ufuk Topcu$^{1}$}
\author{Joydeep Biswas$^{1}$}
\affiliation{%
  \institution{$^{1}$The University of Texas at Austin\quad $^{2}$DEVCOM Army Research Laboratory}
  \country{USA}}
\renewcommand{\shortauthors}{Rana et al.}

\begin{abstract}
Progress in autonomous navigation in off-road environments requires not only advances in algorithms, but also robust and capable physical platforms suited to both development and deployment.
Today, off-road autonomy is typically developed and tested at one of two extremes of vehicle scale, each with different drawbacks.
Full-size vehicles are capable but expensive, require large logistic infrastructure for deployment and maintenance, and expensive when pushed beyond their limits. RC-scale vehicles are small and inexpensive, but limited in the terrain they can traverse and the payload that they can carry, which includes their sensors and compute. We present the AlphaTruck platform, which leverages recent advances in compact sensing and compute to integrate multimodal sensing, embedded CPU-GPU compute, and an onboard autonomy stack for high-speed, off-road navigation, all integrated into a large-scale, 1/5th RC vehicle chassis. AlphaTruck is designed as both a research testbed and a field-deployable platform for commercial and defense missions such as survey, search and rescue, and agriculture, with the capacity to carry task-specific payloads. At 1/5 scale, the autonomy stack, sensors, and batteries all compete for the same mass, power, and cooling budget, so the platform must balance five needs: mobility over rough terrain, rich sensing, onboard computation, payload capacity to carry the autonomy stack while leaving room for task-specific equipment, and practical operation and repair. Thus, AlphaTruck navigates off-road environments autonomously at 6--10~m/s using only onboard sensing and compute.
\end{abstract}


\maketitle
\vspace{-15pt}
\begin{figure}[H]
  \centering
  \includegraphics[width=0.9\linewidth]{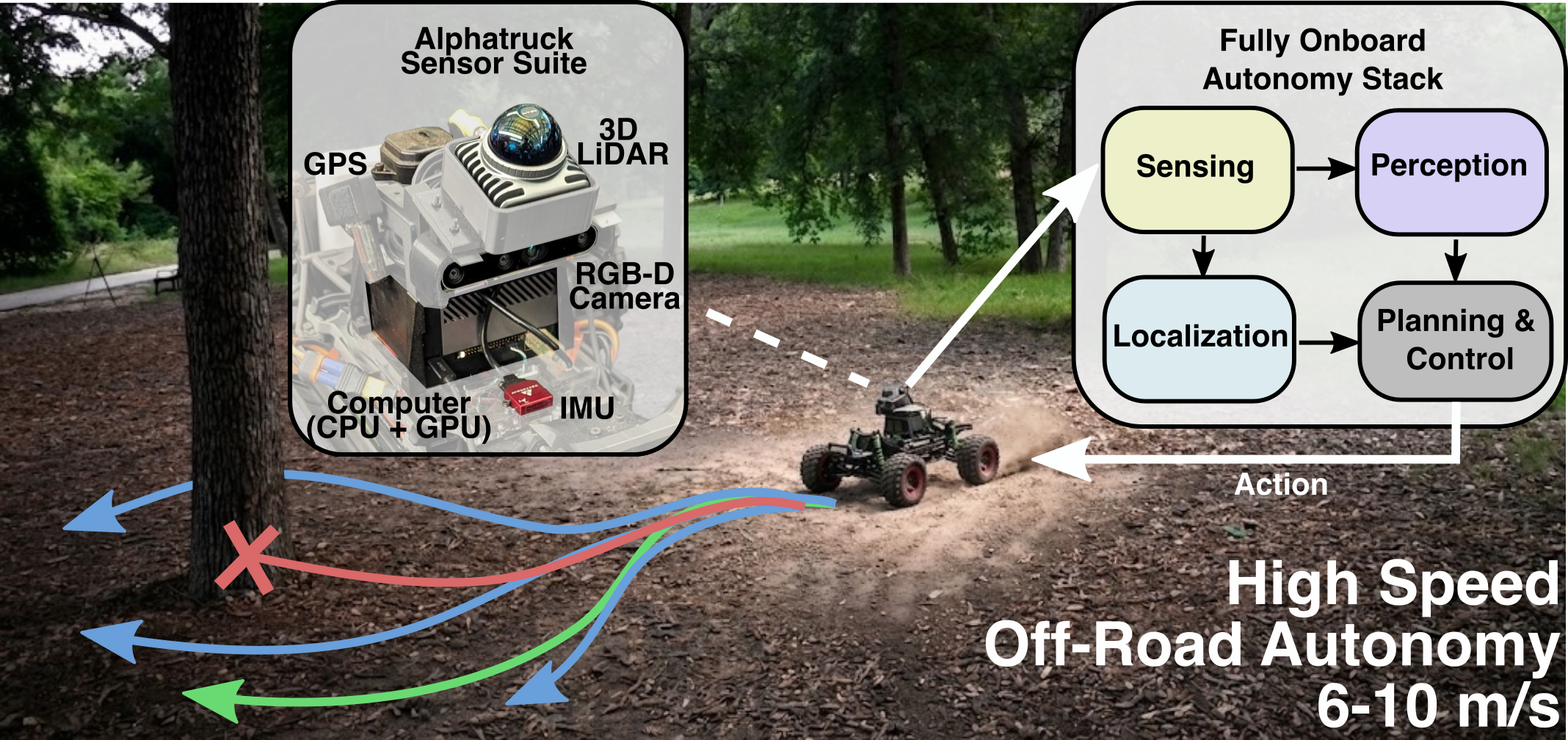}
  \caption{AlphaTruck is a rugged 1/5-scale platform whose multimodal sensors, onboard compute, and full autonomy stack are co-designed for high-speed off-road navigation. In the example shown, the planner evaluates candidate trajectories (blue), rejects a trajectory that collides with the tree (red), and selects the best collision-free trajectory (green) at 6--10~m/s.}
  \label{fig:platform-overview}
\end{figure}

\begingroup
\setlength{\intextsep}{0pt}
\begin{figure}[H]
  \centering

  \begin{subfigure}[t]{0.49\textwidth}
    \centering
    \includegraphics[
      width=\textwidth,
      height=3.0cm,
      keepaspectratio
    ]{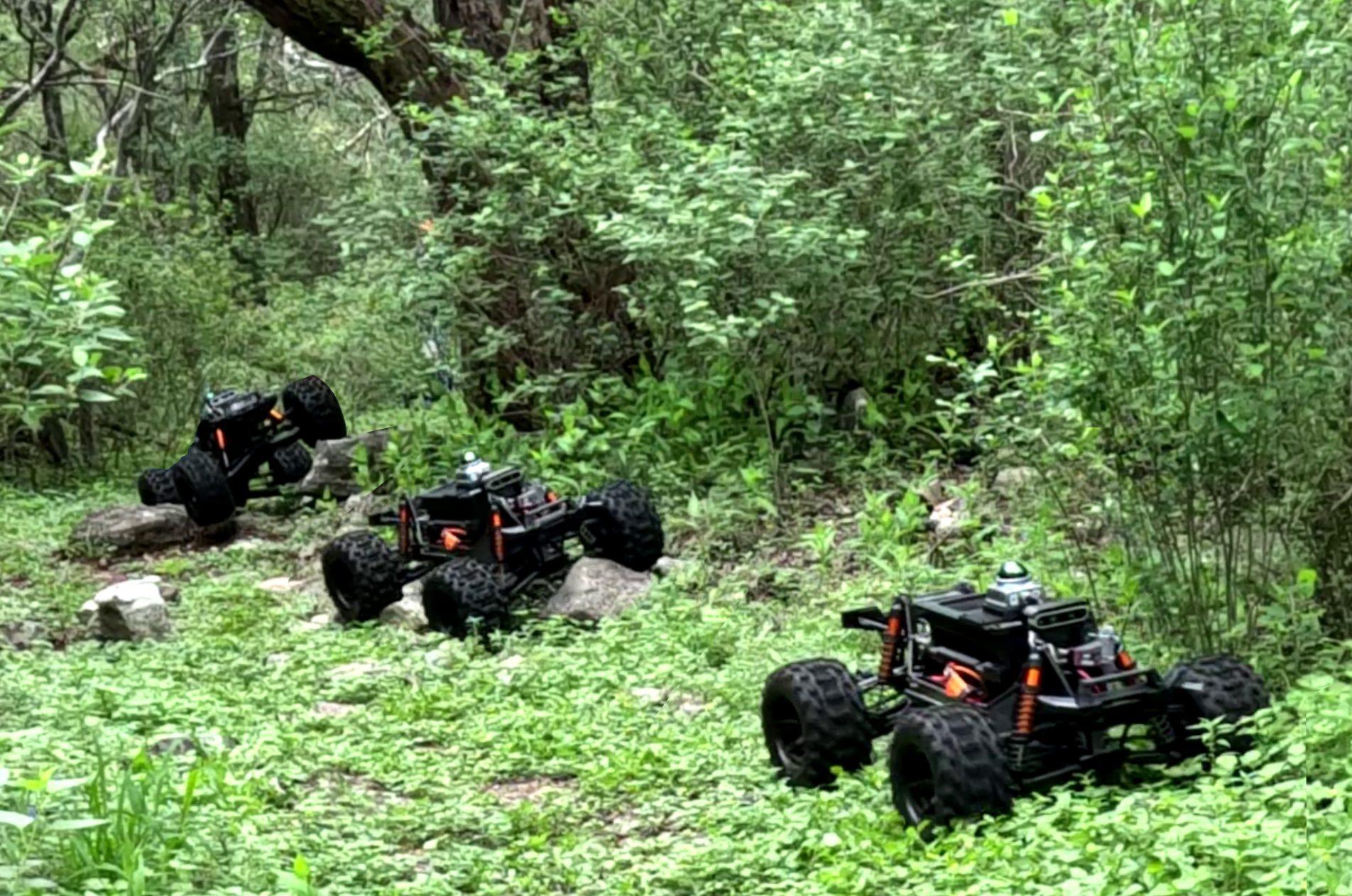}
    \caption{Following a rough trail of rocks, roots, and ruts.}
    \label{fig:hero-trail}
  \end{subfigure}%
  \hspace{0.015\textwidth}%
  \begin{subfigure}[t]{0.49\textwidth}
    \centering
    \includegraphics[
      width=\textwidth,
      height=3.0cm,
      keepaspectratio
    ]{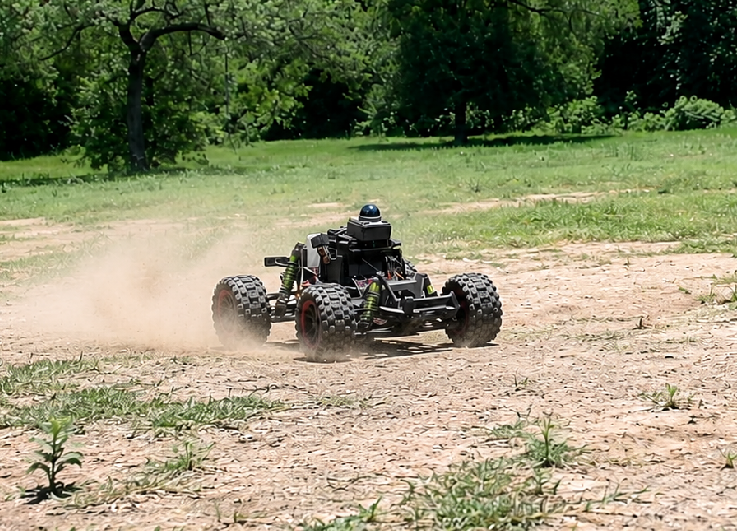}
    \caption{Drifting on open dirt and field.}
    \label{fig:hero-open}
  \end{subfigure}
  \vspace{-4pt} 
  \captionsetup{aboveskip=2pt}
  \caption{ AlphaTruck driving autonomously across unstructured terrain. }
  \label{fig:hero}
  \vspace{-10pt} 
\end{figure}
\endgroup 

\begin{table}[H]
  \captionsetup{skip=2pt}
  \caption{Field-Deployable Platform Comparison. Platforms are compared by speed (high speed: $\geq 5$~m/s), operation on uneven natural terrain, onboard autonomy, range, and cost.}
  \label{tab:platform-comparison}
  \centering
  \scriptsize
  \newcommand{\platformyes}{\textcolor{green!45!black}{\ding{51}}}
  \newcommand{\platformno}{\textcolor{red!70!black}{\ding{55}}}
  \renewcommand{\arraystretch}{0.84}
  \setlength{\tabcolsep}{5pt}
  \begin{tabularx}{\textwidth}{@{}>{\raggedright\arraybackslash}p{0.15\textwidth}>{\raggedright\arraybackslash}X*{5}{>{\centering\arraybackslash}p{0.102\textwidth}}@{}}
    \toprule
    Subclass
      & Platform
      & \shortstack{High\\speed}
      & \shortstack{Natural\\terrain}
      & \shortstack{Full stack\\onboard}
      & \shortstack{km-scale\\range}
      & \shortstack{Under\\\$10k} \\
    \midrule
    \multirow{3}{*}{\shortstack[l]{Small-scale RC\\(1/16--1/10)}}
      & \href{https://arxiv.org/abs/2409.15783}{AnyCar}
      & \platformyes & \platformno & \platformno & \platformno & \platformyes \\
      & \href{https://amrl.cs.utexas.edu/automata.html}{UT Automata}
      & \platformno & \platformno & \platformyes & \platformno & \platformyes \\
      & \href{https://doi.org/10.15607/RSS.2024.XX.021}{HOUND}
      & \platformyes & \platformyes & \platformyes & \platformyes & \platformyes \\
    \cmidrule(lr){1-7}
    \multirow{2}{*}{\shortstack[l]{Intermediate-scale RC\\(1/5)}}
      & \href{https://doi.org/10.1109/MCS.2018.2876958}{AutoRally}
      & \platformyes & \platformno & \platformyes & \platformyes & \platformno \\
      & \textbf{AlphaTruck}
      & \platformyes & \platformyes & \platformyes & \platformyes & \platformyes \\
    \cmidrule(lr){1-7}
    \multirow{2}{*}{\shortstack[l]{Full-size\\drive-by-wire}}
      & \href{https://www.carnegierobotics.com/stallion}{Stallion}
      & \platformyes & \platformyes & \platformyes & \platformyes & \platformno \\
      & \href{https://www.therobotreport.com/oxbotica-pilots-safety-framework-off-road-autonomous-driving/}{Oxbotica Evoque}
      & \platformno & \platformyes & \platformyes & \platformyes & \platformno \\
    \cmidrule(lr){1-7}
    \multirow{3}{*}{\shortstack[l]{Purpose-built\\large UGV}}
      & \href{https://docs.clearpathrobotics.com/docs_robots/legacy/ros1_robots/outdoor_robots/warthog/user_manual_warthog/}{Warthog}
      & \platformno & \platformyes & \platformyes & \platformyes & \platformno \\
      & \href{https://www.nrec.ri.cmu.edu/solutions/defense/off-road-autonomy/}{Crusher}
      & \platformno & \platformyes & \platformyes & \platformyes & \platformno \\
      & \href{https://openreview.net/forum?id=QtVZUPCKrY}{RACER Heavy}
      & \platformyes & \platformyes & \platformyes & \platformyes & \platformno \\
    \bottomrule
  \end{tabularx}
  \vspace{-15pt}
\end{table}
\section{Introduction}

Off-road autonomous driving requires a vehicle to sense, localize, plan, and act on natural terrain, where rocks, roots, ruts, slopes, dirt, and vegetation affect where it can travel and how it responds to steering and throttle. Moreover, GPS and communication are often unreliable in off-road areas, so the full autonomy stack must run onboard. At high speed there is less time to read the terrain ahead and react to bumps or lost traction, so perception, planning, and control must all run quickly.

For this problem, generally two platform sizes have been used (Table~\ref{tab:platform-comparison}). At the large end are drive-by-wire conversions of full-size off-road vehicles and purpose-built UGVs. These platforms carry multimodal sensors and powerful onboard computers and handle rough terrain, but they are costly to transport, support, maintain, and repair, and their mass and speed increase the consequences of a crash, making testing at their limits difficult. At the other extreme, small-scale platforms make experiments inexpensive, portable, and repeatable, but their size limits ground clearance, suspension travel, sensor placement, and onboard computation. After the autonomy hardware is installed, little room may remain for task-specific payloads. Computation can instead be offloaded off the vehicle, but the resulting dependence on a communication link and its latency is unsuitable for long-range missions.

To support rapid experimentation while remaining field-deployable, a platform between these two extremes is needed. In this paper, we present AlphaTruck, a rugged 1/5-scale platform and autonomy stack for high-speed off-road navigation. By \emph{field-deployable}, we mean a platform that can (i)~traverse uneven natural terrain---rocky trails, dirt, roots, ruts, etc.---as shown in Figure~\ref{fig:hero}, (ii)~run its full autonomy stack onboard, (iii)~carry mission-specific payloads beyond that stack, and (iv)~hold enough onboard energy for kilometer-scale missions. Kilometer-scale missions benefit from sustained speed, so we target 6-10~m/s. The autonomy stack limits speed to about 10m/s, beyond which sensor smearing degrades odometry and sensing range limits reaction time. With newer embedded compute and better sensors, AlphaTruck targets unstructured off-road terrain as both a research testbed and a field-deployable platform, driving autonomously off road at 6--10~m/s with the full stack onboard.

\vspace{-8pt}
\begin{figure}[H]
  \centering
  \includegraphics[width=0.8\linewidth]{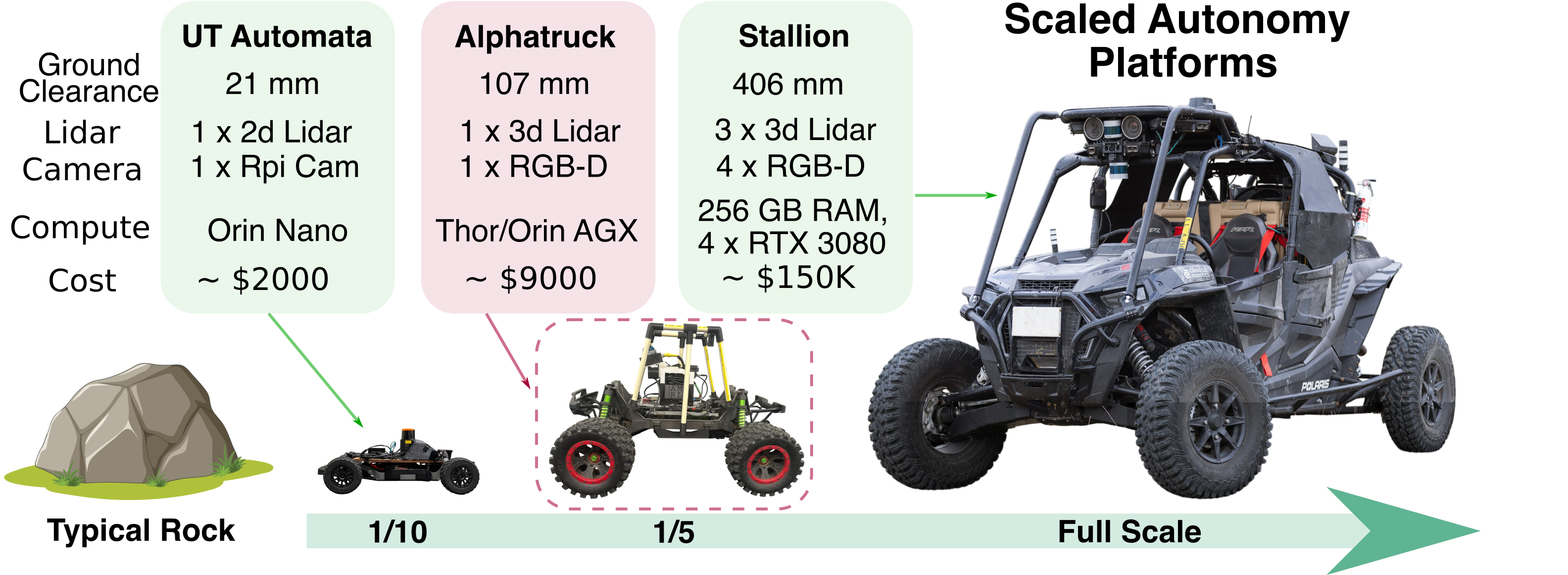}
  \captionsetup{aboveskip=2pt}
  \caption{Comparison of off-road autonomy platforms across vehicle scale, ground clearance, sensing, computation, and cost. AlphaTruck bridges the gap between small research vehicles and full-size platforms.}
  \label{fig:scale}
  \vspace{-14pt}
\end{figure}
\section{Why $1/5^{th}$ Scale?}
AlphaTruck's 1/5 scale balances field deployability with rapid RC experimentation across terrain capability, sensing, onboard computation, payload capacity, and practicality (Fig.~\ref{fig:scale}).
\vspace{-6pt}

\subsection{Terrain capability}
High-speed off-road vehicles often need to traverse rocks, ruts, roots, and small logs rather than avoid them (Fig.~\ref{fig:scale}), requiring sufficient ground clearance and suspension travel. Full-size vehicles have both, but 1/10-scale platforms do not: MuSHR's chassis has only 30~mm of ground clearance without the autonomy hardware~\cite{mushr,redcatBlackoutSC}, so taller rocks and roots hit the chassis and must be driven around. The required capability lies between a 1/10-scale vehicle and a full-size one. Among available 1/5-scale platforms, AlphaTruck uses the Traxxas X-Maxx~\cite{traxxasXMaxx2021} (0.48~m wheelbase); its 107~mm ground clearance and long-travel suspension let it cross rocky trails and obstacles that 1/10-scale platforms often cannot clear (Fig.~\ref{fig:hero-trail}). We modified the stock drivetrain for smooth, controllable motion near zero speed when approaching and crossing obstacles.
\vspace{-8pt}

\subsection{Sensing}
Off-road autonomy requires geometric information about slopes, ditches, rocks, and obstacles, as well as visual information for distinguishing traversable from unsafe terrain. AlphaTruck therefore combines a 3D lidar, an RGB-D camera, an inertial measurement unit (IMU), and an RTK GNSS receiver. The lidar measures terrain geometry; the camera provides appearance and depth; the IMU's accelerometers and gyroscopes measure linear acceleration and rotational motion for localization; and RTK GNSS provides global position when satellite reception is available. This suite captures terrain features that smaller platforms with 2D sensing may miss without requiring the larger sensor suite of a full-size vehicle.
\vspace{-8pt}

\subsection{Onboard computation}
Off-road environments typically lack a reliable link to an external computer, so the full autonomy stack must run concurrently onboard. Full-size platforms meet this with heavy multi-GPU compute: vehicles in this class carry a protected enclosure with a 64-core CPU, 256~GB of RAM, and four RTX~3080 GPUs, drawing roughly 7~kW~\cite{darparacer,roadrunner}. Small platforms make the opposite choice: the UT 1/10-scale RoboRacer uses a Jetson Orin Nano, F1TENTH and MuSHR document Jetson TX2 or Nano boards, and HOUND uses an Orin NX~\cite{utRoboRacer2025,f1tenth,mushr,taliademonstrating}, which leaves little headroom to run 3D-lidar perception, mapping, and sampling-based control together. AlphaTruck runs its entire stack on single embedded Jetson board, avoiding both offboard computation and the multi-GPU setup of a full-size platform. The two variants share the same chassis design and sensor suite, differing only in their compute module: an NVIDIA Jetson AGX Orin or AGX Thor.
\vspace{-10pt}

\subsection{Capacity beyond the autonomy stack}
Field missions require equipment beyond the autonomy stack, such as additional sensors, communication gear, or manipulators, with their own space, mass, and power demands. Full-size vehicles provide this reserve, while small RC platforms often devote nearly all capacity to autonomy hardware and batteries. With its autonomy stack installed, AlphaTruck weighs 11--13~kg and retains space, payload capacity, and energy for mission equipment.
\vspace{-8pt}

\subsection{Practicality}
Full-size vehicles satisfy the first four requirements, but their cost and mass make repeated experimentation difficult: for example, a Stallion with its sensors and stack costs around \$150{,}000, and its size raises transport, maintenance, and repair effort and the consequences of a failure. Small RC vehicles are the opposite, portable, inexpensive to test, and built from replaceable parts that keep the build--test--repair cycle short: MuSHR can be built for about \$900, UT Automata has \$1,500 and \$2,700 configurations, and F1TENTH costs about \$3,600 at 4.5~kg~\cite{mushr,utautomata,goldfain2019autorally}. AlphaTruck costs about \$9000 and keeps these advantages by using a commercial hobby chassis: inexpensive, replaceable parts, one-person transport and operation, and quick repair.
\vspace{-10pt}

\subsection{Convergence at 1/5 scale}
AutoRally showed that 1/5-scale vehicles, which weighed nearly 22~kg~\cite{goldfain2019autorally}, could support aggressive onboard autonomy for rally racing. New compact CPU--GPU computers and 3D sensors let AlphaTruck retain this scale while leaving capacity for payloads and longer missions. Its limited onboard computation and control-cycle time motivate the design principles in the next section.
\vspace{-8pt}

\section{The AlphaTruck Platform}
We now describe AlphaTruck, the 1/5-scale platform we have developed for high-speed off-road autonomy (Figure~\ref{fig:horns}). It combines an off-road vehicle with onboard sensing, computation, and a complete autonomy stack, and its design follows three principles:
\vspace{-3pt}

\begin{figure}[t]
  \centering
  \includegraphics[width=\linewidth]{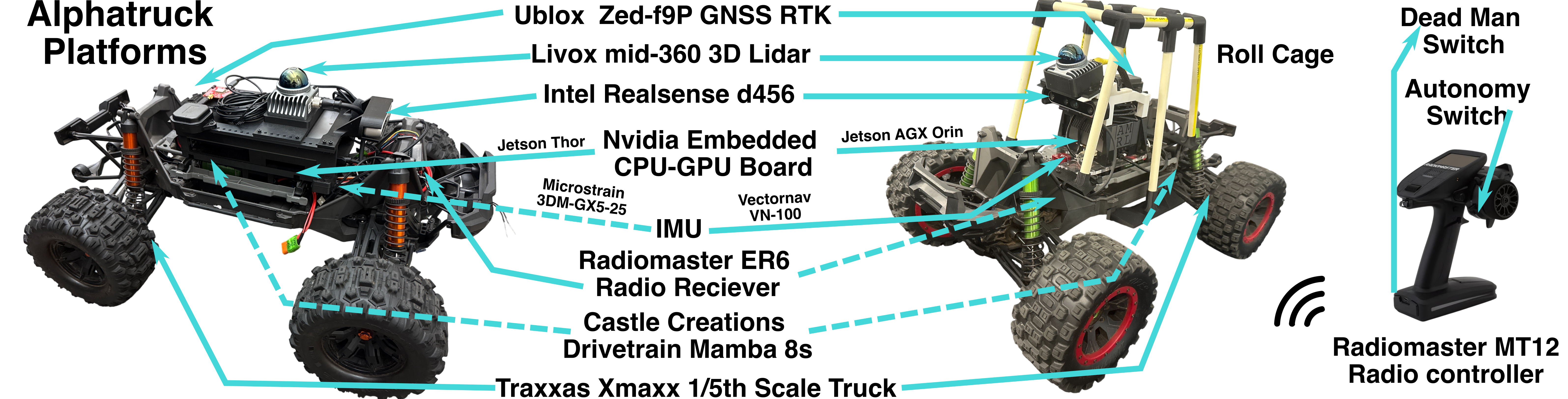}
  \caption{AlphaTruck's integrated sensing, computation, communication, and safety hardware. The components are mounted on a rugged 1/5-scale chassis for high-speed off-road operation.}
  \label{fig:horns}
\end{figure}

\begin{enumerate}[label=(\roman*),leftmargin=1.8em]
  \item \textbf{Platform-aware, fault-tolerant hardware design.} Sensors and compute are placed to balance visibility, stability, and protection. Breakaway mounts and replaceable parts limit damage from impacts and rollovers and simplify repair.
  \item \textbf{Fidelity--latency tradeoff-guided algorithm selection.} A model's performance in the control loop matters more than its accuracy alone. A higher-fidelity model helps only if the loop can still sense and replan fast enough.
  \item \textbf{System-level co-design.} Localization, perception, mapping, planning, and control share sensor preprocessing and intermediate results instead of maintaining separate pipelines that repeat the same work.
\end{enumerate}
The following subsections show how these principles shape the hardware, autonomy stack, software, simulation, and failure recovery.

\subsection{Hardware and Chassis}
\textbf{Chassis and drivetrain.} The commercial chassis described in \S2 uses a Castle Mamba Monster~X 8S electronic speed controller (ESC) and 1717-1260Kv sensored motor in place of the stock sensorless drivetrain~\cite{castleMambaMonsterX8S,castle1717Motor}. Rotor-position feedback allows the ESC to commutate cleanly near zero speed, enabling precise low-speed PWM control for creeping over obstacles while retaining high-speed operation. The ESC also reports voltage, current, RPM, and temperature telemetry. The onboard computer sends steering and throttle over a binary serial link with separate command and telemetry packets, protected by a cyclic redundancy check (CRC)~\cite{williams1993crc}, and commands are clipped to safe ranges. An ExpressLRS (ELRS) transmitter~\cite{expresslrsFAQ} provides autonomy-enable and deadman switches for manual override, and the speed controller cuts motor power if the radio link is lost.

\textbf{Sensing and compute.} The 3D lidar is a Livox Mid-360, the RGB-D camera is an Intel RealSense D456, the IMU is either a MicroStrain 3DM-GX5-25 or a VectorNav VN-100, and the RTK GNSS receiver is a u-blox ZED-F9P. A single embedded compute module processes all sensor data. The camera is mounted high for lookahead, and the lidar is pitched about $14.3^\circ$ forward and downward to cover the terrain ahead. A roll cage and breakaway mounts protect them during rollovers, while the GNSS antenna sits on a conductive ground plane for reliable reception. Two 4S LiPo batteries power traction, sensing, and computation.

\vspace{-10pt}
\subsection{Autonomy Stack}

\begin{figure}[t]
  \centering
  \includegraphics[width=0.7\linewidth]{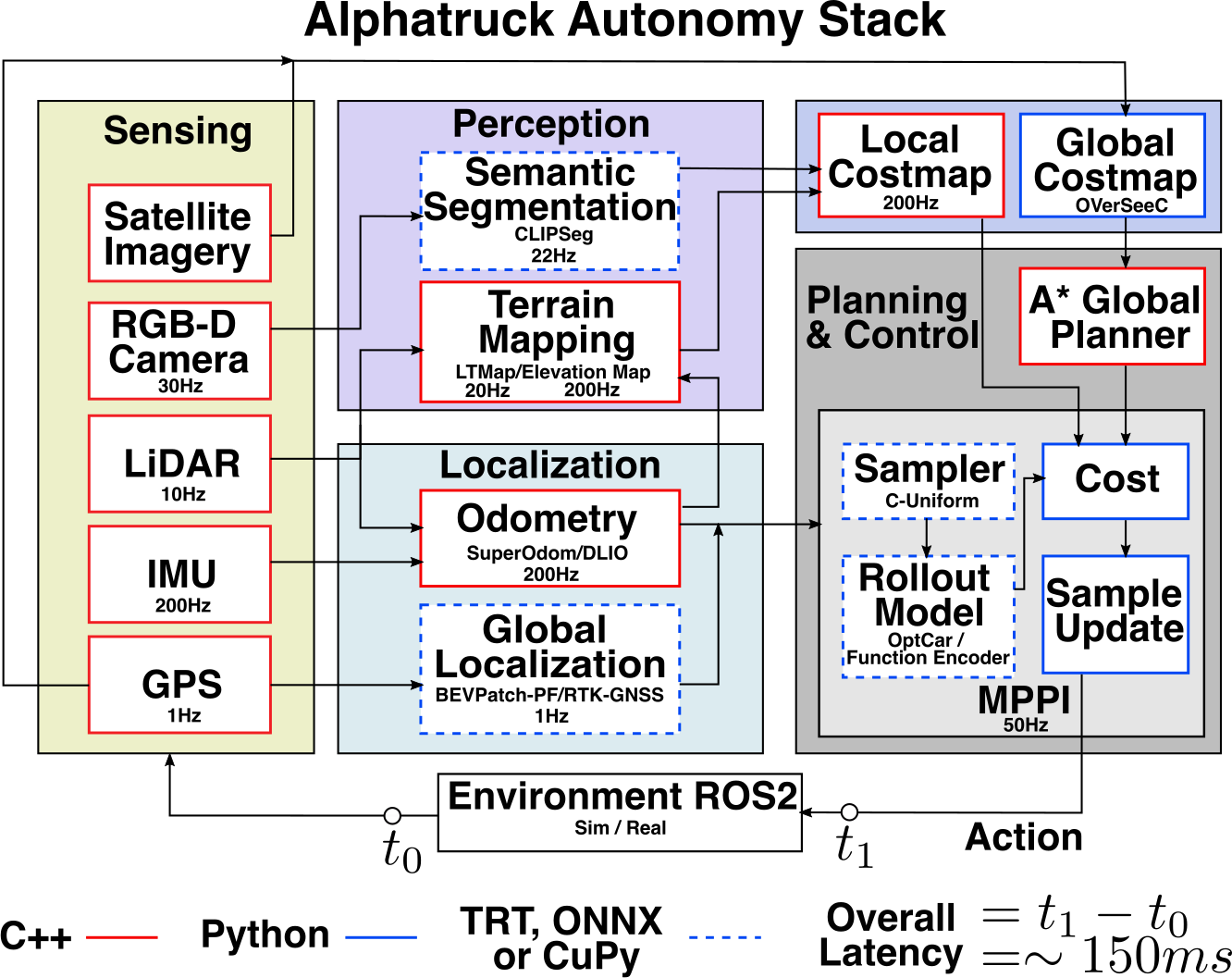}
  \caption{This diagram shows how sensor data flows through AlphaTruck's ROS~2 autonomy stack. ROS~2 enables the same software to run in simulation and on the vehicle. Learned methods use Python, ONNX/TensorRT for model optimization and quantization, and CuPy for custom CUDA kernels; other modules use C++ for speed.}
  \label{fig:autonomy-stack}
\end{figure}

We now describe the autonomy software stack currently deployed on AlphaTruck. The ROS~2 stack in Figure~\ref{fig:autonomy-stack} uses Python or C++ based on the workload: learned models use Python with ONNX/TensorRT for optimized inference, CuPy runs custom CUDA kernels for further acceleration, and most non-learned modules use C++ for low-latency execution. Localization, perception, and planning and control run concurrently on a single embedded GPU.

\textbf{Localization.} The planner needs an estimate of the vehicle's position and orientation, called its pose, on every control cycle. AlphaTruck estimates this pose using lidar-inertial odometry (LIO), with either DLIO~\cite{chen2023dlio} or SuperOdom~\cite{zhao2021superodom}. LIO combines changes between successive 3D lidar scans with acceleration and rotational-motion measurements from the 200~Hz IMU to produce a pose estimate at approximately 200~Hz. It does not depend on camera images, which can blur during fast motion. Unlike full simultaneous localization and mapping (SLAM), which also builds and globally corrects a map, LIO focuses on providing the fast local motion estimate needed by the controller. AlphaTruck handles global correction separately by matching against a prior map with KISS-ICP~\cite{vizzo2023kissicp} when one exists, or by using RTK GNSS otherwise. We do not use wheel encoders because wheel slip makes them less informative at high speed.

\textbf{Perception and terrain mapping.} The camera identifies vegetation, surface types, and terrain beyond the lidar range, while the lidar measures elevation and obstacles. The individual perception models are off-the-shelf; AlphaTruck's contribution is to co-design them into one shared pipeline rather than run each separately. Elevation mapping is embedded in the LIO module, reusing its tilt correction, filtering, and lidar stream while a separate thread builds the elevation map. The map is built at the 10~Hz lidar rate and shifted and repositioned at the 200~Hz odometry rate. Semantic traversability comes from ClipSeg~\cite{lueddecke22_cvpr} at 24~Hz, fused with RealSense depth into a bird's-eye view. We alternatively use the lightweight terrain map (LTMap)~\cite{ltmap}, which accumulates RGB point clouds into a rolling 2.5D grid of elevation, color, and height-variation layers with optional learned cost, aging out old observations rather than keeping them permanently.

\textbf{Planning and control.} Planning and control turn the current state and terrain estimate into a feasible action within the control period. AlphaTruck uses MPPI ~\cite{williams2018mppi} in a modular implementation whose sampler, rollout model, cost, and action-update rule can each be replaced independently, running on the GPU with PyTorch and CuPy and converting learned modules to TensorRT. Two design choices manage this cost. The first is the sampler. Standard MPPI perturbs a nominal control sequence with Gaussian noise, so its rollouts cluster near that nominal trajectory and explore few alternatives. The deployed C-Free-Uniform sampler~\cite{cao2025c} instead conditions on the local map and spreads samples uniformly through free space, giving a better plan for the same compute without changing the controller. Second, AlphaTruck rolls out a learned forward model of one of two kinds, a generalist specialized offline to this platform~\cite{rana2026adaptinggeneralistvehiclemodels}, or an online model in which function encoders adapt a compact dynamics representation from streaming data within the loop~\cite{ward2026zero}. Either model must meet the MPPI deadline alongside localization and perception, competing for the same GPU. In the deployed configuration, MPPI with 600 samples, a 2~s horizon, and a 0.02~s step runs at 50~Hz with a bicycle-model rollout and 30~Hz with a $\sim$600k-parameter learned model, all with localization, mapping, and perception running concurrently on the same GPU.
\vspace{-10pt}

\subsection{Software Infrastructure (workspace, repos \& driver management)}
The stack must run across two compute variants whose host operating systems, library versions, and accelerator configurations differ: the Jetson AGX Orin and Jetson AGX Thor. It therefore encapsulates all dependencies in a single container. A custom Docker wrapper (\texttt{./container}) is the sole entry point: it mounts the source, exposes host networking, and forwards devices (GPU, cameras, serial buses), reading environment-specific parameters (device mounts, ports, vehicle addresses, container names) from a \texttt{config/} directory to build a fully parameterized invocation per host, with utility tasks under \texttt{scripts/}. Process startup is declared in tmuxinator manifests that bring up the ROS~2 daemon, sensor drivers, and control interfaces in one command.
\vspace{-10pt}

\subsection{Simulation and Sim-to-Real Parity}
Because the autonomy stack is modular, the same navigation code runs unchanged against any environment, whether a simulator or the real car in the field, selected by a single backend flag. The simulation backends span a range of fidelity: the analytical kinematic and dynamic bicycle models we built are fast, deterministic, and closed-loop for controller and planner development, while Isaac Lab~\cite{mittal2023orbit} and BeamNG.tech~\cite{beamng_tech}, two existing simulators we integrate, add GPU-parallel sensor rendering for batched learning and realistic off-road physics, respectively. An idea can thus be developed against a bicycle model, validated in BeamNG, and deployed on hardware. We still validate on hardware because off-road tire--terrain interactions vary substantially with surface and speed. Reproducing them would require tire, suspension, terrain-deformation, and friction models whose parameters are hard to measure and vary by site: dirt and vegetation change tire slip, rocks and ruts change wheel loading, and the vehicle then brakes or turns differently than predicted.
\vspace{-10pt}

\section{What AlphaTruck Enables}

\begin{figure}[t]
  \centering
  \includegraphics[width=0.8\linewidth]{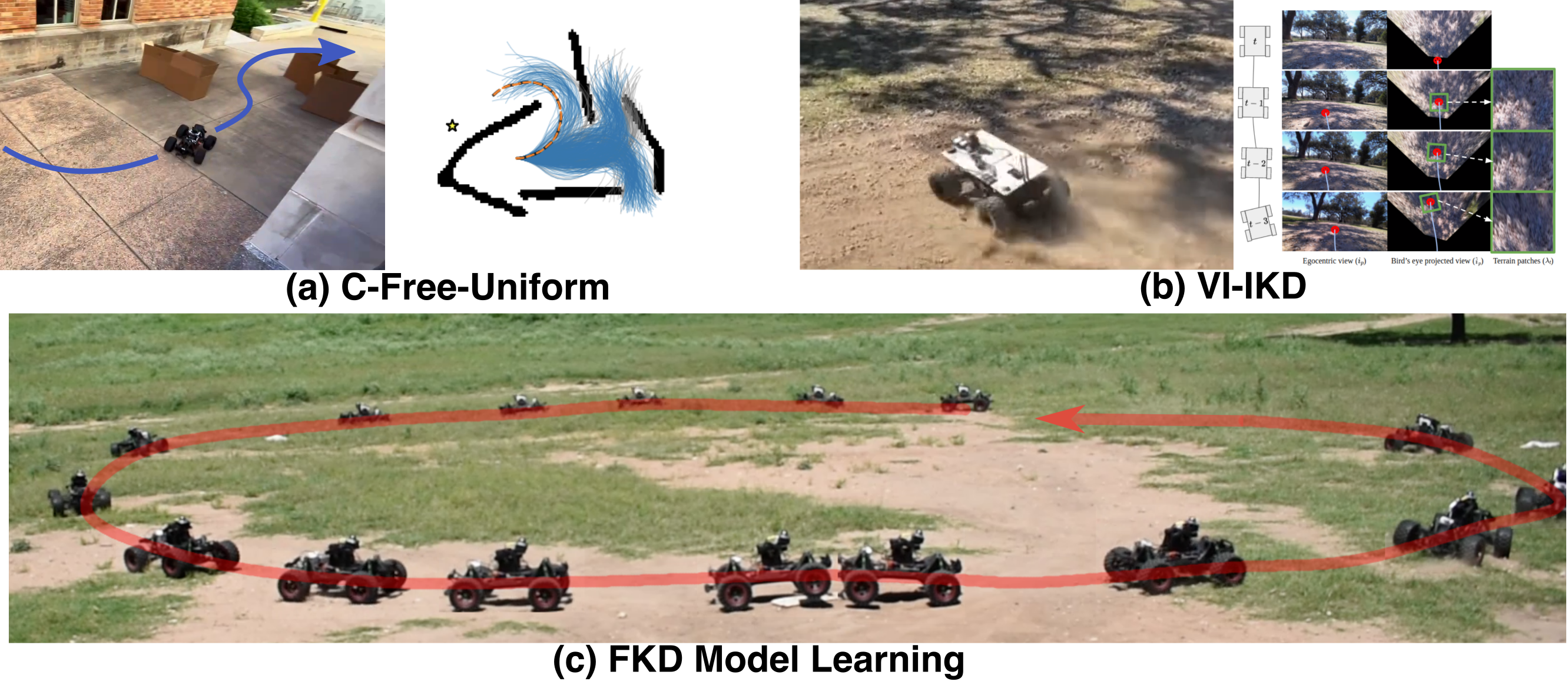}
  \captionsetup{aboveskip=2pt}
  \caption{Capabilities demonstrated on AlphaTruck: (a)~C-Free-Uniform sampling-based planning~\cite{cao2025c}; (b)~visual-inertial inverse kinodynamics (VI-IKD)~\cite{viikd}; and (c)~forward-kinodynamic (FKD) model learning with OptCar~\cite{rana2026adaptinggeneralistvehiclemodels} and online adaptation with function encoders~\cite{ward2025terrain,ward2026zero}.}
  \label{fig:alphatruck-enables}
\end{figure}

AlphaTruck brings together research capabilities developed independently on earlier platforms (Figure~\ref{fig:alphatruck-enables}). We trace how each research line led to AlphaTruck and what the platform enables next.
\vspace{-10pt}

\subsection{Learned Kinodynamics}
Model-based planning and control depend on accurate vehicle dynamics. Analytical bicycle models suffice at low speed, but at high speed, terrain-dependent slip, drift, and ground reactions dominate. VI-IKD combines inertial history with a forward-looking visual patch to anticipate terrain-dependent slip~\cite{viikd}. OptCar takes a complementary approach by specializing a generalist forward model to the target vehicle~\cite{rana2026adaptinggeneralistvehiclemodels}. AlphaTruck's camera and IMU provide the inputs needed by VI-IKD, while its onboard GPU allows OptCar to serve as the MPC rollout model.
\vspace{-10pt}
\subsection{Online Dynamics Adaptation}
As vehicle speed increases, changes in terrain cause larger changes in slip and steering response, so a fixed dynamics model quickly loses accuracy. Our function-encoder approach adapts a compact vehicle-dynamics representation from streaming odometry without retraining~\cite{ward2025terrain,ward2026zero}. MPPI uses the adapted model to predict how the vehicle will respond on the current terrain. On AlphaTruck, this adaptation runs while localization, perception, and planning operate concurrently onboard.
\vspace{-4pt}

\subsection{Sampling-Based Planning and Control}

More MPPI samples improve planning, but the control deadline limits each cycle, making the sampling strategy critical at high speed. Sample efficiency matters most off road, where obstacles are irregular and a 2~s horizon at 10~m/s spans 20~m. C-Uniform spreads trajectories across the reachable configuration space~\cite{poyrazoglu2025c}, and its unsupervised extension supports longer horizons~\cite{poyrazoglu2025unsupervised}. On AlphaTruck, C-Free-Uniform uses the local terrain map to focus samples in collision-free space~\cite{cao2025c}, while uncertainty-guided exploration helps escape local minima in clutter~\cite{ugempc}.
\vspace{-10pt}

\subsection{Terrain Perception, Mapping, and Localization}
Traversal costs depend on the mission: a routine inspection may prefer a dirt trail over grass, while a time-critical mission accepts grass for a shorter route. Rather than hand-editing terrain costs or retraining the model, the user supplies this preference at deployment. PACER and SPACER learn terrain costmaps from example-based preferences and generalize them to unseen terrain~\cite{pacer,spacer}. For global planning from satellite imagery, OVerSeeC accepts mission preferences in natural language~\cite{oversee}. Context-adaptive planner parameters~\cite{appld} and the lightweight terrain-mapping backbone the stack builds on~\cite{ltmap} extend the same line. More recent work grounds mapless navigation in foundation-model features with learned rewards~\cite{creste}, while BEV-Patch-PF geo-localizes without GPS by matching onboard features to aerial imagery~\cite{bevpatchpf}. These methods connect directly to the platform's constraints: speed makes a costmap error a crash rather than a detour, so AlphaTruck tests these costmaps closed-loop at consequential speed, and BEV-Patch-PF is the fallback for the GPS-denied condition that shaped the deployed localization design. Finally, AlphaTruck's robustness and hardware can help take our group's prior energy-aware and multi-robot mission planning, from energy-optimal paths and solar-aware planning to data-mule and UAV--UGV scheduling~\cite{energy_dubins, plonski2013solar, bhadauria2011datamules, uav_ugv_ag, air2ground}, off road at speed.
\vspace{-10pt}

\section{Lessons from Building and Operating AlphaTruck}

Building and repeatedly testing AlphaTruck showed that failures can originate in autonomy algorithms, hardware, software, timing, or the operating environment, and often involve interactions among them. Routine repairs and restarts reduced field-testing time, so we treated repairability and field procedures as part of the platform design. Model training, simulation, and onboard deployment differ in their throughput, graphics, power, and timing requirements. We therefore use compute resources suited to each stage.

\vspace{-10pt}

\subsection{Development Workflow and Compute Resources}
Model development and evaluation span several compute environments, each serving a different stage of the workflow.

\begin{itemize}[leftmargin=1.2em]
  \item We have a three stage workflow. First, offline training uses large parallel simulation runs for reinforcement learning, simulated data collection, model training, and hyperparameter sweeps. Next, each model is tested in closed-loop simulation as part of the full navigation stack (Fig.~\ref{fig:autonomy-stack}), exposing model errors, latency, and models that are too large for real-time execution. We then revise or retrain the model, adjust accuracy--latency tradeoffs, or optimize inference with ONNX, TensorRT, and custom CUDA kernels. Finally, we deploy the navigation stack on AlphaTruck for missions and data collection, which feeds the next training cycle.
  \item We use three GPU classes across the workflow. Data-center GPUs such as H100, A100, A40, and GB200 systems provide the tensor throughput, memory, and scaling used for training and large parallel workloads. GPUs with ray-tracing hardware, such as A40, RTX PRO 6000, and consumer RTX cards, support closed-loop simulation and rendering. Jetson AGX Orin and Thor run the navigation stack onboard under power, thermal, and real-time constraints. These roles overlap, particularly between offboard training and simulation.
\end{itemize}
\vspace{-10pt}

\subsection{Software Architecture for Development and Deployment}
Development needs frequent code changes; deployment needs none.

\begin{itemize}[leftmargin=1.2em]
  \item \textbf{Development.} Because we change and test autonomy algorithms frequently, the stack is modular and hooks into different environments through a single interface. The same navigation code runs in simulation and on the vehicle, with only the environment behind that interface differing. This shortens control-algorithm debugging and lets an algorithm move from simulation to hardware without modification.
  \item \textbf{Deployment.} The less code an operator touches during a deployment, the better. Debugging still needs a terminal; running an experiment should not. We built a lightweight laptop and phone app for stack bringup, and a deployment interface should go further: reporting sensor health and per-topic data rates, and displaying the vehicle's live position on a satellite or prerecorded map on which the operator places mission waypoints.
\end{itemize}

\vspace{-6pt}
\subsection{Real-Time Onboard Compute Budget}
Everything from sensing to the next command must finish within a fixed time budget on every cycle, and the onboard compute available for it is limited. Staying inside that budget takes both algorithmic choices, in how the controller and models are computed, and operational care in what else is allowed to run on the vehicle.

\begin{itemize}[leftmargin=1.2em]
  \item Each stage of the sensing, estimation, and planning pipeline takes time to compute, so the command applied on a given cycle acts on state that is already stale by the pipeline's processing time. A controller that ignores this latency tracks poorly at high speed; explicit compensation substantially improves circular trajectory tracking (Figure~\ref{fig:deployment-lessons}(a)). Low-pass filtering the sampled MPPI action is a low-cost measure that further improves closed-loop tracking.
  \item For sampling-based MPC such as MPPI, which rolls out the dynamics model hundreds of times per cycle, a higher-fidelity learned model helps mainly at high speed; below approximately 4 m/s, an analytical bicycle model tracks as well, rolls out faster, and needs no training.
  \item Online adaptation must also fit within the fixed control-cycle budget. In our function-encoder model, the neural basis functions remain fixed while a small set of coefficients represents the current vehicle dynamics. We estimate these coefficients from streaming odometry using a least-squares update rather than gradient-based retraining, leaving more of each cycle for MPPI trajectory rollouts~\cite{ward2025terrain,ward2026zero}.
  \item Visualization tools such as RViz consume enough compute and bandwidth to lower the stack's loop rate, so we disable them during runs and use them only for offline debugging.

\end{itemize}
\vspace{-10pt}

\subsection{Hardware Design and Maintainability}
The hardware is chosen and modified to withstand rough field use and stay quick to service.
\begin{figure}[t]
  \centering
  \includegraphics[width=0.8\linewidth]{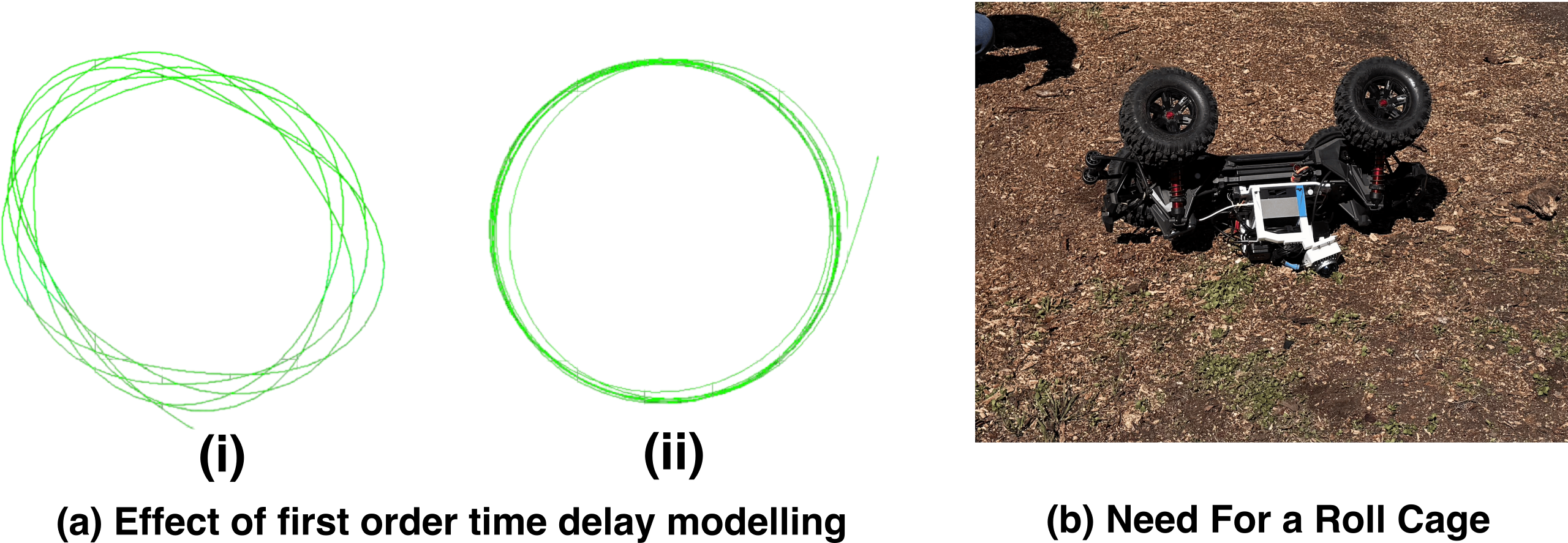}
  \caption{Lessons from field deployment. (a)~Repeated circular trajectories (i)~without and (ii)~with latency compensation show that accounting for command latency substantially improves real-world tracking. (b)~A rollover during an aggressive maneuver damaged exposed sensors before a protective roll cage was installed.}
  \label{fig:deployment-lessons}
\end{figure}

\begin{itemize}[leftmargin=1.2em]
  \item High-torque maneuvers and tight spaces require smooth motion at very low speed. The stock sensorless ESC lacks rotor-position feedback and cannot commutate cleanly near zero speed. We replaced it with a Castle Mamba Monster~X 8S ESC and 1717-1260Kv sensored motor~\cite{castleMambaMonsterX8S,castle1717Motor}; the rotor feedback enables precise low-speed PWM control while retaining high-speed operation.
  \item High-speed off-road operation causes frequent impacts and rollovers; one rollover damaged exposed sensors before the roll cage was installed (Figure~\ref{fig:deployment-lessons}(b)). The cage now protects the sensors, while sacrificial 3D-printed mounts break first and can be replaced quickly. The commercial chassis makes most parts inexpensive and readily available---a stripped spur gear was sourced locally within two hours. On the other hand the lidar, which costs roughly \$1{,}000, accounts for over a tenth of the platform cost and cannot be sourced locally, so spares have to be budgeted and ordered well ahead of a deployment.
  \item An operator must be able to stop the vehicle even when the software or the radio link fails, so the emergency stop sits on dedicated hardware rather than the operator's computer and cuts motor power the moment the radio link drops.
  \item High ambient temperature throttles the GPU enough to slow inference and cause navigation failures, overheats sensors, and in one deployment disabled the onboard network router and cut the communication link. A shaded, actively cooled compute enclosure would mitigate this; full-size testbeds such as RACER and AutoRally already protect their electronics with such an enclosure~\cite{darparacer,goldfain2019autorally}, and adding one remains an outlook item for this platform.
\end{itemize}
\vspace{-10pt}
\subsection{Instrumentation and Failure Diagnosis}
Much of the development effort goes to diagnosing failed runs rather than improving autonomy algorithms, so the platform includes extensive instrumentation to identify their causes.

\begin{itemize}[leftmargin=1.2em]
  \item A sensor-health interface detects corrupted data and low update rates before they disrupt a run. Live and logged voltage, current, motor speed, temperature, and link quality help trace apparent autonomy failures to thermal, power, or communication faults.

  \item We monitor odometry by overlaying the estimated trajectory and RTK GNSS position on a satellite map. RTK GNSS provides this reference with little computation and also supplies ground truth for post-deployment training and evaluation.
  \item Model-based control depends on the accuracy of its dynamics model, whether analytical at low speed or learned at higher speed. A standard workflow that replays a recorded log and overlays each model's predicted rollout against the observed trajectory allows a model fault to be localized quickly and the appropriate response, whether parameter adjustment for an analytical model or retraining for a learned model, to be selected.
  \item Recording all topics overloads memory and compute on the embedded modules and has caused sensor drivers to fail, so only the topics an experiment requires are recorded, and which topics those are is set by a per-experiment configuration file rather than specified by hand on each run.
  \item Image-based perception is sensitive to illumination: direct sunlight entering the lens saturates the image and requires reduced white balance and contrast, whereas overcast conditions require the opposite. A mechanism to adjust these parameters rapidly for the prevailing conditions, and preferably an adaptive scheme, is needed to keep perception reliable.
\end{itemize}
\vspace{-10pt}
\section{Conclusion}
We presented AlphaTruck, a 1/5-scale platform for high-speed autonomy on unstructured terrain. It is large enough to carry 3D sensing, onboard GPU computation, batteries, and mission payloads, yet small and inexpensive enough to transport, repair, and test repeatedly. In our deployments, the vehicle ran its full autonomy stack onboard, including localization, terrain mapping, planning, analytical or learned dynamics, and control, and drove autonomously at 6-10~m/s. AlphaTruck provides a common platform for testing learned kinodynamics, online adaptation, terrain-aware planning, and future long-range missions under realistic field conditions. Repeated deployments showed that reliable off-road autonomy depends not only on algorithms, but also on latency compensation, compute and thermal management, safety hardware, logging, and repairability. Overall, AlphaTruck fills the practical gap between small laboratory platforms and full-size off-road vehicles, providing a capable and accessible platform for continued research and field deployment.

\bibliographystyle{ACM-Reference-Format}
\bibliography{bibliography}


\end{document}